\documentclass[10pt,peerreview,draftclsnofoot,onecolumn,a4paper]{IEEEtran}

\usepackage{amsmath} 
\usepackage{charter}
\usepackage{microtype,color}

\def\x{\boldsymbol{x}}
\def\y{\boldsymbol{y}}
\def\a{\boldsymbol{a}}
\def\u{\boldsymbol{u}}
\def\I{\mathbf{I}}
\def\P{\mathbf{P}}

\newtheorem{theorem}{Theorem}

\newtheorem{proposition}[theorem]{Proposition}

\begin{document}

\title{On the convergence of the IRLS algorithm in Non-Local Patch Regression}

\author{Kunal N. Chaudhury$^{*}$ \thanks{ $^{*}$K. N. Chaudhury is with the Program in Applied and Computational Mathematics (PACM), Princeton University, Princeton, NJ 08544, USA (mail: kchaudhu@math.princeton.edu).}
}

\maketitle

\begin{abstract}
Recently, it was demonstrated in \cite{Chaudhury2012,Chaudhury2013} that the robustness of the classical Non-Local Means (NLM) algorithm \cite{BCM2005} can be improved by incorporating $\ell^p (0 < p \leq 2)$ regression into the
NLM framework. This general optimization framework, called Non-Local Patch Regression (NLPR), contains NLM as a special case. Denoising results on synthetic and natural images show that NLPR  consistently performs better than NLM 
beyond a moderate noise level, and significantly so when $p$ is close to zero. An iteratively reweighted least-squares  (IRLS) algorithm was proposed for solving the regression problem in NLPR, where the NLM output was used to initialize the iterations. Based on 
exhaustive numerical experiments, we observe that  the IRLS algorithm is globally convergent  (for arbitrary initialization) in the convex regime  $1 \leq  p \leq 2$, and locally convergent (fails very 
rarely using NLM initialization) in the non-convex regime $0 < p < 1$. In this letter, we adapt the ``majorize-minimize'' framework introduced in \cite{Voss1980} to explain these observations. 
\end{abstract}

\begin{keywords}
Non-local means, non-local patch regression, $\ell^p$ minimization, non-convex optimization, iteratively reweighted least-squares, majorize-minimize, stationary point, relaxation sequence, linear convergence. 
\end{keywords}

\section{Introduction}

In the last decade, some very effective frameworks for image restoration have been proposed that (a) exploit long-distance correlations in natural images, and (b) use patches instead of pixels to robustly compare
photometric similarities. This includes the classical Non-Local Means (NLM) algorithm \cite{BCM2005}, and the more sophisticated BM3D algorithm \cite{BM3D}. The latter combines the NLM framework with other classical algorithms, 
and is widely considered as the state-of-the-art in image 
denoising. We refer the reader to \cite{Milanfar2013} for a comprehensive review of patch-based algorithms.

Let $u = (u_i)$ be some linear indexing of the input noisy image. In NLM, the restored image $\hat{u} = (\hat{u}_i)$ is computed using the simple formula
\begin{equation}
\label{NLM_formula}
\hat{u}_i  = \frac{\sum_{j \in S(i)} w_{ij} u_j}{\sum_{j \in S(i)} w_{ij} }.
\end{equation}
Here, $w_{ij}$ is some weight (affinity) assigned to pixels $i$ and $j$, and $S(i)$ is some non-local (sufficiently large) neighborhood of pixel $i$ over which the averaging is performed \cite{BCM2005}. In particular, for a given pixel $i$, let $\P_i$ denote the restriction of $u$ to a square window around $i$. Letting $k$ be the length of this window, this associates every pixel $i$ with a point $\P_i$ in $\mathbf{R}^{k^2}$ (the patch space). The weights in NLM are set to be $
 w_{ij} = \exp(- \lVert \P_i - \P_j \rVert^2 / h^2)$, where $\lVert \P_i - \P_j \rVert$ is the Euclidean distance between $\P_i$ and $\P_j$, and $h$ is a smoothing parameter. 

Recently, it was demonstrated in \cite{Chaudhury2012,Chaudhury2013} that the robustness of the Non-Local Means (NLM) algorithm \cite{BCM2005} can be improved by incorporating $\ell^p$ regression into the NLM framework. The idea was to fix some $0 < p \leq 2$, and consider the following unconstrained optimization on the patch space:
\begin{equation}
\label{Lp_regression}
\hat{\P}_i =  \arg \ \min_{\P} \  \sum_{j \in S(i)} w_{ij} \  \lVert \P - \P_j \rVert^p,
\end{equation}
where $w_{ij}$ are the weights used in NLM (one could also use other weights, e.g., see \cite{Sun2013}). The denoised pixel $\hat{u}_i$ was then set to be the center pixel in $\hat{\P}_i$. Note that this reduces to the
simple formula in \eqref{NLM_formula} when $p=2$. In this case, the optimization is performed pixel-wise. For any other value of $p$, 
the optimization in \eqref{Lp_regression} becomes a generic optimization on the patch space -- the regression needs to be performed on patches and not just pixels. In particular, when $0 < p \leq 1$, the resulting estimator turns
out to be more robust to ``outliers'' in the patch space (compared to standard NLM), and this leads to significant improvement in the denoising quality. We refer the reader to \cite{Chaudhury2013} for an intuitive 
understanding of the robustness in NLPR, and for denoising results on synthetic and natural images.

Note that we can generally write the optimization problem in \eqref{Lp_regression} as
\begin{equation}
\label{cost}
\min_{\x \in \mathbf{R}^d}  \ \sum_{j=1}^n w_j \lVert \x - \a_j \rVert^p,
\end{equation}
where $\a_1,\ldots,\a_n$ are given points in $\mathbf{R}^d$, and $w_1,\ldots,w_n$ are some positive weights. Motivated by the work on algorithms for $\ell^p$ minimization in \cite{Chartrand2008,DDFG2009}, the authors 
in  \cite{Chaudhury2013} proposed to optimize \eqref{cost} using iteratively reweighted least-squares (IRLS). Fixing some small $ \varepsilon > 0$, and initializing $\x^{(0)}$ as the NLM output, the update rule was set to be
\begin{equation}
\label{FP}
\x^{(t+1)} = \frac{\sum_j \mu^{(t)}_j \a_j}{\sum_j \mu^{(t)}_j} \qquad (t \geq 0),
\end{equation}
where
\begin{equation}
\label{weights}
\mu^{(t)}_j = w_j (|| \x^{(t)} - \a_j ||^2 + \varepsilon)^{p/2 -1}.
\end{equation}
We refer the reader to \cite{Chaudhury2013} for the heuristics behind the IRLS update in \eqref{FP}. Extensive numerical experiments show us the that the  algorithm is globally convergent  (for arbitrary $\x^{(0)}$) in the 
convex regime  $1 \leq  p \leq 2$, and locally convergent (fails very rarely compared to, say, the gradient or Newton method) in the non-convex regime $0 < p < 1$, provided that $x^{(0)}$ is set as the NLM estimate. The 
former observation can be explained using the existing literature on the Weiszfeld algorithm \cite{Weiszfeld1937,Voss1980}, which is perhaps less well-known in the signal processing community. In this letter, we adapt 
the ``majorize-minimize'' framework introduced in \cite{Voss1980} to specifically analyze \eqref{FP} when $0 < p < 1$. The analysis automatically covers 
the case $1 \leq p \leq 2$. This is the content of Section \ref{Analysis}. In particular, we will show that the algorithm forces the cost to be non-increasing as the iteration progresses, and that 
it exhibits linear convergence both in the convex and non-convex (when convergence does occur) regime. 
\textcolor{black}{We note that in \cite{DDFG2009}, the authors have done an analysis of a related (but more complex) IRLS algorithm in the non-convex regime, but the analysis is quite involved. The present analysis is much more simple, 
and is based on elementary results from smooth unconstrained optimization.}

\section{Convergence Analysis}
\label{Analysis}

The key question is what is the cost associated with the IRLS iterations in \eqref{FP}? This must be resolved even before we ask questions about convergence. It turns out that the iterations 
corresponds to a regularized version of 
the original cost \eqref{cost}. This is given by
\begin{equation}
\label{Weber}
\Phi_{\varepsilon}(x)  =  \sum_{j=1}^n w_j \lvert \x - \a_j \rvert_{\varepsilon}^p,
\end{equation}
where $\lvert \x \rvert_{\varepsilon}$ is the regularized version of the Euclidean norm, 
\begin{equation*}
\lvert \x \rvert_{\varepsilon}^2 = \lVert \x \rVert^2 + \varepsilon \qquad (\varepsilon > 0).
\end{equation*}
Note that $\Phi_{\varepsilon}$ corresponds to the original cost \eqref{cost} when $\varepsilon = 0$. It can be argued that the minimizers of $\Phi_{\varepsilon}$ converge to the minimizer of the original problem as 
$\varepsilon$ tends to zero. In other words, for sufficiently small $\varepsilon$, the minimizer of $\Phi_{\varepsilon}$ is close to that of  the original problem. Henceforth, to simplify notation, we fix some small $\varepsilon > 0$ and denote $\Phi_{\varepsilon}$ by $\Phi$. We note that in \cite{Chaudhury2013}, the authors proposed to start with, say,  $\varepsilon = 1$, and then gradually shrink it to zero as the iteration progresses. While this 
does tend to speed up the convergence, the associated analysis becomes quite complicated.

The advantage we get by considering the regularized problem is that the function $\Phi(x)$ is smooth (infinitely differentiable). This allows us to use the powerful tools of smooth
optimization. Moreover, $\Phi$ inherits the convex nature of the original problem, namely, that it is strictly convex for $1 \leq p \leq 2$. Since $\Phi$ is smooth, it 
suffices to show that its Hessian is positive definite. In fact, the gradient (denoted by $\Phi'$) and the Hessian (denoted by $\Phi''$) are given by 
\begin{equation*}
\Phi'(\x) =  p \sum_j w_j \lvert \x - \a_j \rvert_{\varepsilon}^{p-2} (\x-\a_j),
\end{equation*}
and
\begin{equation*}
\Phi''(\x) =  p\sum_j w_j  \lvert \x - \a_j \rvert_{\varepsilon}^{p-4} \Big[ \lvert \x - \a_j \rvert_{\varepsilon}^2 \I_d - (2-p) (\x-\a_j)(\x-\a_j)^T \Big].
\end{equation*}
Here, $\I_d$ is the identity matrix of size $d \times d$. For any non-zero $\u \in \mathbf{R}^d$, 
\begin{equation*}
\u^T \Phi''(x) \u = p \sum_j w_j  \lvert \x - \a_j \rvert_{\varepsilon}^{p-4} \Big[  \lvert \x - \a_j \rvert_{\varepsilon}^2 \lVert \u \rVert^2 - (2-p) (\u^T(\x-\a_j))^2 \Big].
\end{equation*}
Since $\u^T (\x - \a_j) \leq \lVert \u \rVert \cdot \lVert \x - \a_j \rVert <  \lVert \u \rVert \cdot \lvert \x - \a_j \rvert_{\varepsilon}$, the quadratic form is strictly larger than
\begin{equation*}
p(p-1) \lVert \u \rVert^2  \sum_j w_j \lvert \x - \a_j \rvert_{\varepsilon}^{p-2} .
\end{equation*}
This is non-negative if and only if $1 \leq p \leq 2$. Therefore, $\Phi$ is strictly convex in this case, and has a unique global minimizer $\x^{\star}$ for which $\Phi'(\x^{\star}) = 0$. 
On the other hand, it is not difficult to see that $\Phi$ need not be convex when $p <1$. The best we can hope for in this case is that the iterates in \eqref{FP} converge to some local stationary point. In fact, we can show
that 

\begin{theorem}
\label{Thm1} The IRLS update in \eqref{FP} guarantees the following:

\begin{enumerate}

\item For $0 \leq p \leq 2$, the sequence $(\Phi(\x^{(t)}))$ is strictly monotonic, $\Phi(\x^{(t+1)}) \leq \Phi(\x^{(t)})$ for all $t$. 
 
\item  When $1 \leq p \leq 2$,  $(\x^{(t)})$ converges linearly to the unique global minimizer of $\Phi$.

\item  For $0 < p < 1$, under some additional assumptions, the convergence is again linear and the limit of $(\x^{(t)})$ is a stationary point of $\Phi$.

\end{enumerate}

\end{theorem}

By linear convergence, we mean that the convergence happens at an exponential rate. The relaxation property is particularly important for the non-convex setting, providing the guarantee that the cost at the end of the iterations is 
less than that obtained from the initial estimate $\x^{(0)}$.
In optimization literature, one calls $(\x^{(t)})$ a \textit{relaxation sequence} if  $\Phi(\x^{(t+1)}) \leq \Phi(\x^{(t)})$. Since $\Phi$ is trivially bounded below, this implies convergence of the sequence $(\Phi(\x^{(t)}))$. As 
we will see, the iterates in \eqref{FP} \textit{unconditionally} generate a relaxation sequence in both the convex and non-convex regime. This property turns out to be a central ingredient in the 
guarantees provided by Theorem \ref{Thm1}. \textcolor{black}{We note that the relaxation property was recently observed in \cite{Sun2013} for the special case $p=1$. However, we remark that the fact that the convergence of $(\Phi(\x^{(t)}))$ is 
not sufficient to guarantee convergence of $(\x^{(t)})$, as claimed in \cite{Sun2013}. For example, it is possible that $(\x^{(t)})$ keeps oscillating, or escapes to infinity, while ensuring that $\Phi(\x^{(t+1)}) \leq \Phi(\x^{(t)})$. }One of the cases can 
however be ruled out immediately:
\begin{proposition}
\label{bdd}
For $0 < p \leq 2$, the IRLS iterates $(\x^{(t)})$ are bounded (do not escape to infinity). 
\end{proposition}

This is a simple consequence of the observation that $\Phi(\x) \rightarrow \infty$ as $\lVert \x \rVert \rightarrow \infty$. Indeed, if $(\x^{(t)})$ does escape to infinity, then we would have a contradiction since we 
just showed that $(\Phi(\x^{(t)}))$ is bounded. 

In the convex regime, we will show that oscillations can also be ruled out. To do so in the non-convex regime, we will need additional assumptions.

\subsection{Majorize-Minimize interpretation}

To establish Theorem \ref{Thm1}, we will use the majorize-minimize (MM) framework from \cite{Voss1980}. In this framework, the key idea is to globally approximate $\Phi$ from above using a sequence of quadratic functions. 
More precisely, after having found $\x^{(t)}$, we construct a function $\Psi_t(\x) = \Psi(\x; \x^{(t)})$ such that

\begin{itemize}

\item $\Phi(\x) \leq \Psi_t(\x)$ for all $x$.

\item $\Phi(\x^{(t)}) = \Psi_t(\x^{(t)})$.

\item $\Psi_t(\x)$ has $\x^{(t+1)}$ in \eqref{FP} as its unique global minimizer.

\end{itemize}

Once we have $\Psi_t$ with the above properties, we immediately see that
\begin{equation*}
\Phi(\x^{(t+1)}) \leq \Psi_t(\x^{(t+1)}) \leq \Psi_t(\x^{(t)}) = \Phi(\x^{(t)}). 
\end{equation*}
That is, we are guaranteed that $( \x^{(t)})$ is a relaxation sequence. We now need to specify $\Psi_t(\x)$.

\begin{proposition}
The following choice will suffice:
\begin{equation}
\label{Quad}
\Psi_t(\x) = \Phi(\x^{(t)}) + (\x - \x^{(t)})^T \Phi'(\x^{(t)}) + \frac{p}{2} \sum_j \mu_j^{(t)} \lVert \x - \x^{(t)} \rVert^2,
\end{equation}
where $\mu^{(t)}_j$ is as defined in \eqref{weights}.
\end{proposition}

Note that the linear part of $\Psi_t(\x)$ is simply the linear approximation of $\Phi$ at $\x^{(t)}$, and the quadratic form is derived from the dominant part of $\Phi''(\x^{(t)})$. 
By construction, $\Psi_t(\x^{(t)}) = \Phi(\x^{(t)})$. Moreover, 
it is clear that $\Psi_t(x)$ is strictly convex (for all $p$), and has a global unique minimizer. Setting $x^{(t+1)}$ to be this minimizer, we have 
\begin{equation}
\label{SP}
\Psi_t'(\x^{(t+1)}) = \Phi'(\x^{(t)}) + p \sum_j \mu_j^{(t)} (\x^{(t+1)} - \x^{(t)}) = 0.
\end{equation}
Substituting for $\Phi'$, we get the update rule in \eqref{FP}.

To complete the proof, we need to show that $\Phi(\x) \leq \Psi_t(\x)$. Note that we can write $\Phi(\x) - \Psi_t(\x)$ as
\begin{eqnarray*}
\sum_{j=1}^n w_j \Big( \lvert \x^{(t)} - \a_j \rvert_{\varepsilon}^p  - \lvert \x - \a_j \rvert_{\varepsilon}^p  &+ & p \sum_j w_j  \lvert \x^{(t)} - \a_j \rvert_{\varepsilon}^{p-2} (\x-\x^{(t)})^T(\x^{(t)}-\a_j) \\
&+& \frac{p}{2} \sum_j \lvert \x^{(t)} - \a_j \rvert_{\varepsilon}^{p-2}  \lVert \x - \x^{(t)} \rVert^2\Big). 
\end{eqnarray*}
We substitute the following above:
\begin{equation*}
(\x - \x^{(t)})^T(\x^{(t)} - \a_j)  =  (\x - \a_j)^T(\x^{(t)} - \a_j) - \lVert \x^{(t)}- \a_j \rVert^2,
\end{equation*}
and
\begin{equation*}
 \lVert \x - \x^{(t)} \rVert^2 = \lVert \x - \a_j \rVert^2 + \lVert \x^{(t)} - \a_j  \rVert^2 - 2(\x-\a_j)^T(\x^{(t)}-\a_j).
\end{equation*}
This allows us to simplify the expression to
\begin{eqnarray*}
\sum_{j=1}^n w_j \Big(\alpha_j^{p/2} - \beta_j^{p/2}  +   \frac{p}{2} \alpha_j^{p/2 - 1} (\beta_j - \alpha_j) \Big). 
\end{eqnarray*}
where $\alpha_j= \lvert  \x^{(t)} - \a_j \rvert_{\varepsilon}^2$ and $\beta_j = \lvert \x - \a_j \rvert_{\varepsilon}^2$. It can be verified that each term in the sum is non-negative for 
any $0 \leq p \leq 2$ (for $p=0,1,$ and $2$ this is obvious). This shows that $\Phi(x) \leq \Psi_t(x)$, concluding the proof of \eqref{Quad}.

\subsection{Global and local convergence}

Since the sequence $(\Phi(x^{(t)}))$ is monotonic and bounded below, it is convergent. In particular, $\Phi(x^{(t)}) - \Phi(x^{(t+1)}) \rightarrow 0$ as $t \ \rightarrow \infty$.
So, what can we say about the sequence $(\x^{(t)})$?

\begin{proposition}
\label{diff}
We claim that $\lVert \x^{(t)} - \x^{(t+1)} \rVert  \rightarrow 0$ as $t$ gets large.
\end{proposition}

To do so, we use \eqref{SP},
\begin{equation*}
\Phi'(\x^{(t)}) = - p \sum_j \mu_j^{(t)} (\x^{(t+1)} - \x^{(t)}),
\end{equation*}
and the majorizing property,
\begin{equation*}
\Phi(\x^{(t+1)}) \leq \Psi(\x^{(t+1)}) = \Phi(\x^{(t)}) + (\x^{(t+1)} - \x^{(t)})^T \Phi'(\x^{(t)}) + \frac{p}{2} \sum_j \mu_j^{(t)} \lVert \x^{(t+1)} - \x^{(t)} \rVert^2.
\end{equation*}
Combining these, we see that
\begin{equation*}
 \Phi(\x^{(t+1)}) \leq  \Phi(\x^{(t)}) -  \frac{p}{2} \sum_j \mu_j^{(t)} \lVert \x^{(t+1)} - \x^{(t)} \rVert^2.
\end{equation*}
Now, from \eqref{weights} we have the trivial bound $\mu_j^{(t)}  >  w_j \varepsilon^{p/2-1}$. We can then write
\begin{equation*}
 \lVert \x^{(t+1)} - \x^{(t)} \rVert^2 \leq   \gamma \big[ \Phi(\x^{(t)}) - \Phi(\x^{(t+1)}) \big],
\end{equation*}
where 
\begin{equation*}
\gamma=\frac{2\varepsilon^{1-p/2}}{p (\sum_j w_j)}.
\end{equation*}
Since  $(\Phi(x^{(t)}))$ is convergent, we arrive at our claim.

Note that cannot directly conclude from Proposition \ref{diff} that $(\x^{(t)})$ is convergent.  
However, since $(\x^{(t)})$ is bounded, it has convergent subsequences (by compactness). In the convex regime, we can say something more:

\begin{proposition}
For $1 \leq p \leq 2$, every convergent subsequence has the same limit, and this limit is the unique stationary point of $\Phi$. In particular, 
it is necessary that $(\x^{(t)})$ converges to $\x^{\star}$
\end{proposition}

We now establish the above claim. Let $(\x^{(m)})$ be a subsequences that converges to $\x^{\star}$. We know that 
$\lVert \x^{(m)} - \x^{(m+1)} \rVert  \rightarrow 0$, so that the limit of both $(\x^{(m)})$ and $(\x^{(m+1)})$ must be $\x^{\star}$. Note that
\begin{equation*}
 0 = \Psi'(\x^{(m+1)}) = \Phi'(\x^{(m)}) +  p \sum_j \mu_j^{(m)} (\x^{(m+1)} - \x^{(m)}).
\end{equation*}
Since both $\Phi'(\x)$ and $\mu_j=\mu_j(\x)$ are smooth, letting $m \rightarrow \infty$, we have
\begin{equation*}
0 = \Phi'(x^{\star}) +  p \sum_j w_j \lvert \x^{\star} - \a_j \rvert^{p-2} (\x^{\star} - \x^{\star}) =  \Phi'(\x^{\star}) .
\end{equation*}
That is, $\x^{\star}$ is a stationary point of $\Phi$. In the convex regime $1 \leq p \leq 2$, we know that $\Phi$ has a unique stationary point $\x^{\star}$. Therefore, every convergent subsequence 
of $(\x^{(t)})$ must have the same limit $\x^{\star}$. 

\begin{proposition}
For $0 < p < 1$, the above claim is true only under certain local assumptions. 
\end{proposition}

The problem in this case is that there can be multiple stationary points of $\Phi$, and the previous argument breaks down (as is typical with non-convex problems). 
All we know in this case is that $(\x^{(t)})$ is bounded, and that the entire $(\x^{(t)})$ can be restricted to a ball 
$B_r(\x^{\star})$ of radius $r$ around $\x^{\star}$. Suppose we assume that that the initialization $\x^{(0)}$ is ``good'', in that it is situated sufficiently close to a local (probably global) minimizer $\x^{\star}$. 
It is then possible that $r$ is small enough and $B_r(\x^{\star})$ contains no other stationary points of $\Phi$. In this case, we are guaranteed that every convergent subsequence, and hence the whole sequence $(\x^{(t)})$, converges to $\x^{\star}$.

\subsection{Convergence rate}

Plots of $\log(\Phi(\x^{(t+1)}) - \Phi(\x^{\star}))$ versus $t$ for \eqref{FP} suggests a linear trend both for the convex and non-convex cases (assuming convergence for the latter). For the convex regime, we 
can indeed guarantee that

\begin{proposition}
\label{linconv}
For $1 \leq p \leq 2$, 
\begin{equation*}
\Phi(\x^{(t+1)}) - \Phi(\x^{\star}) \leq \nu \big( \Phi(\x^{(t)}) - \Phi(\x^{\star}) \big) \qquad (\text{large  } t, \ \nu < 1). 
\end{equation*}
\end{proposition}

 In other words, the convergence is exponential, where the convergence rate is controlled by $\nu$. To establish our claim, we being by comparing $\Phi$ at the points $\x^{(t+1)}$ and the 
 linear combination $\theta_t \x^{(t)} + (1-\theta_t)\x^{\star}$ (we will define $\theta_t$ later),
\begin{equation*}
 \Phi(\x^{(t+1)}) \leq \Psi_t(\x^{(t+1)}) \leq \Psi_t(\theta_t \x^{(t)} + (1-\theta_t)\x^{\star}).
\end{equation*}
The first inequality follows from majorization, and the second from the optimality of $\x^{(t+1)}$. From \eqref{Quad}, we can write $\Psi_t(\theta_t \x^{(t)} + (1-\theta_t) \x^{\star})$ as
\begin{equation*}
\Phi(\x^{(t)})  + (1-\theta_t) (\x^{\star} - \x^{(t)})^T \Phi'(\x^{(t)}) + \frac{p(1-\theta_t)^2}{2}  \lVert \x^{\star} -\x^{(t)} \rVert^2  \sum_j \mu_j^{(t)}.
\end{equation*}
On the other hand,
\begin{equation}
\label{opt}
\Psi_t(\x^{\star}) = \Phi(\x^{(t)}) + (\x^{\star} - \x^{(t)})^T \Phi'(\x^{(t)}) + \frac{p}{2} \lVert \x^{\star} - \x^{(t)} \rVert^2 \sum_j \mu_j^{(t)}.
\end{equation}
Using this to eliminate the term containing $\Phi'(\x^{(t)})$, we can write $\Psi_t(\theta_t \x^{(t)} + (1-\theta_t) \x^{\star})$ as
\begin{eqnarray*}
\theta_t \Phi(\x^{(t)})  + (1-\theta_t)\Big( \Psi(\x^{\star}) - \frac{p\theta_t}{2} \sum_j \mu_j^{(t)} \lVert \x^{\star} - \x^{(t)} \rVert^2 \Big).
\end{eqnarray*}
Now, set
\begin{equation}
\label{bound}
 \theta_t = \frac{2(\Psi(\x^{\star}) - \Phi(\x^{\star}))}{p \lVert \x^{\star} - \x^{(t)} \rVert^2 \sum_j \mu_j^{(t)}}.
\end{equation}
Then $ \Phi(\x^{(t+1)}) \leq \Psi_t(\x^{(t+1)}) = \theta_t \Phi(\x^{(t)})  + (1-\theta_t)\Psi(\x^{\star})$, and hence
\begin{equation*}
 \Phi(\x^{(t+1)}) - \Phi(\x^{\star}) \leq \theta_t (\Phi(\x^{(t)}) - \Phi(\x^{\star})).
\end{equation*}
By construction, $\theta_t \geq 0$ for all $t$. We are done if we can show that $\theta_t \leq \nu <1$ for some constant $\nu$. By Taylor's theorem,
\begin{equation}
\label{Taylor}
\Phi(\x^{\star}) =  \Phi(\x^{(t)}) + (\x^{\star} - \x^{(t)})^T \Phi'(\x^{(t)}) + \frac{1}{2} (\x^{\star} - \x^{(t)})^T \Phi''(\y^{(t)}) (\x^{\star} - \x^{(t)}),
\end{equation}
where $\y^{(t)}$ is some point on the segment joining $\x^{\star}$ and $\x^{(t)}$. Plugging \eqref{opt} and \eqref{Taylor} in \eqref{bound}, 
\begin{equation*}
 \theta_t = 1 - \frac{(\x^{\star} - \x^{(t)})^T \Phi''(\y^{(t)}) (\x^{\star} - \x^{(t)})}{p \lVert \x^{\star} - \x^{(t)} \rVert^2 \sum_j \mu_j^{(t)} } \leq 1 - \Big(p \sum_j  \mu_j^{(t)} \Big)^{-1} \lambda_{\min}( \Phi''(\y^{(t)}) ),
\end{equation*}
where $\lambda_{\min}(A)$ denotes the smallest eigenvalue of the matrix $A$. Now, since $\Phi''$ is continuous, $\lambda_{\min}( \Phi''(\y^{(t)}))$ approaches $\lambda_{\min}(\Phi''(\x^{\star}))$ as $t \rightarrow \infty$.
Moreover, $\x^{\star}$ is a (local) minimizer. Hence, $\Phi''(\x^{\star})$ is necessarily positive semidefinite, that is, $\lambda_{\min}(\Phi''(\x^{\star})) \geq 0$. This is true for $0 < p \leq 2$.  
Therefore, for all sufficiently large $t$, $0 \leq \theta_t \leq \nu \leq 1$, where 
\begin{equation*}
 \nu = 1 - \Big(p \sum_j  \mu_j^{(\infty)} \Big)^{-1} \lambda_{\min}(\Phi''(\x^{\star})),
\end{equation*}
and where $\mu^{(\infty)}_j = w_j (|| \x^{\star} - \a_j ||^2 + \varepsilon)^{p/2 -1} > 0$.

For the convex regime $1 \leq p \leq 2$, $\Phi''(\x^{\star})$ is guaranteed to be positive definite, 
so that $\nu <1$. This completes the proof of Proposition \ref{linconv}. 

For the non-convex setting, the above argument holds under additional assumptions:

\begin{proposition}
 For $0 < p <1$, assume that $(\x^{(t)})$ converges to the local minimizer $\x^{\star}$, and that $\Phi''(\x^{\star})$ is positive definite. Then $(\Phi(\x^{(t)}))$ converges linearly to $\Phi(\x^{\star})$.
\end{proposition}

\section{Discussion}
\label{Discussion}

\textcolor{black}{We note that it is rather difficult to extend the analysis in \cite{Sun2013} to the non-convex setting $0<p<1$. Moreover, it is not possible to estimate the convergence rate using the analysis in \cite{Sun2013}. What their paper shows is that the cost is non-increasing for the case $p=1$. One cannot study the behavior of the iterates using this result alone. For example, as our analysis shows, while the cost is non-increasing when $0<p<1$, it is not sufficient to guarantee convergence. We also note that the technique of proof used in \cite{Weiszfeld1937} and \cite{Voss1980} are quite different from that used is our paper. The basic idea of ``majorize-minimize'' is the same, but the mathematical ideas used in the present paper are different. Voss, for example, uses the sophisticated KKT theory (on polytopes) in his analysis. As against this, our analysis uses elementary ideas from smooth unconstrained optimization. Moreover, it is not obvious how the analysis in \cite{Weiszfeld1937} and \cite{Voss1980} can be extended to the non-convex regime $0<p<1$.}

The linear convergence of IRLS is typical of first-order methods. We have also tried second-order Newton methods for optimizing \eqref{Weber}, which are guaranteed
to exhibit quadratic convergence (locally). Indeed, Newton methods typically require less than half the number of iterations needed by the updates in \eqref{FP} to reach a given accuracy.
However, the cost of a single Newton step (computation of $\Phi''$ and its inversion) is significantly more than that of the simple update in \eqref{FP}. As a result, the total execution time of IRLS turns 
out to be smaller than that of Newton methods. The other point that we noticed from the numerical simulations is that IRLS is much more stable than Newton (or gradient descent) methods in the non-convex regime. 
In particular, the iterates in the Newton method often diverge to infinity if the initialization is not ``close'' to a local minimum. However, the IRLS iterates never escape to 
infinity (this is clear from Theorem \ref{Thm1}), and almost always converge to the global optimum when we initialize using the NLM output. In rare cases when \eqref{FP} gets ``stuck'' in a local minimum (say, 
due to bad initialization), Newton methods are found to have the same problem.  It would be interesting to see if one could give a more accurate analysis of the IRLS algorithm in the non-convex regime.

\textcolor{black}{Finally, we note that the analysis in this letter are about the convergence properties of the NLPR algorithm in \cite{Chaudhury2013}, and not about its denoising performance. These are two entirely different aspects of the algorithm. In  \cite{Chaudhury2013}, empirical results about the performance of NLPR were reported, but the issue of convergence was not studied. We are currently investigating the ``optimality'' of the denoising performance of NLPR and possible scope for improvements, which will be reported in a future correspondence}. 

\section{Acknowledgments}
The author would like to thank Prof. Gilad Lerman and Prof. Amit Singer for interesting discussions.

\end{document}